\documentclass[10pt,twocolumn,letterpaper]{article}

\usepackage{cvpr}              %

\usepackage[dvipsnames]{xcolor}

\definecolor{cvprblue}{rgb}{0.21,0.49,0.74}
\usepackage[pagebackref,breaklinks,colorlinks,citecolor=cvprblue]{hyperref}
\usepackage{multirow}
\usepackage{tabularx}
\usepackage{makecell}
\usepackage{array}
\usepackage{caption}
\usepackage{float}
\def\paperID{6802} %
\def\confName{CVPR}
\def\confYear{2024}

\title{Towards Robust and Accurate Visual Prompting}

\author{Qi Li$^{1}$ \qquad Liangzhi Li$^1$\thanks{Corresponding author.} \qquad Zhouqiang Jiang$^1$ \qquad Bowen Wang$^2$\\
$^1$MeetYou AI Lab $^2$Osaka University\\
{\tt\small liqi@u.nus.edu, liliangzhi@xiaoyouzi.com, jiangzhouqiang93@gmail.com, wang@ids.osaka-u.ac.jp}
}

\begin{document}
\maketitle
\begin{abstract}
Visual prompting, an efficient method for transfer learning, has shown its potential in vision tasks. However, previous works focus exclusively on VP from standard source models, it is still unknown how it performs under the scenario of a robust source model: Whether a visual prompt derived from a robust model can inherit the robustness while suffering from the generalization performance decline, albeit for a downstream dataset that is different from the source dataset? In this work, we get an affirmative answer of the above question and give an explanation on the visual representation level. Moreover, we introduce a novel technique named Prompt Boundary Loose (PBL) to effectively mitigates the suboptimal results of visual prompt on standard accuracy without losing (or even significantly improving) its adversarial robustness when using a robust model as source model. Extensive experiments across various datasets show that our findings are universal and demonstrate the significant benefits of our proposed method.
\end{abstract}
\vspace{-0.2in}
    
\section{Introduction}
\label{sec:intro}
Harnessing knowledge from large-scale models allows for efficient training on new tasks compared to learning from scratch \cite{pan2009survey,raina2007self,chen2021exploring,bao2021beit,he2022masked}. Researchers have proposed numerous paradigms for conducting knowledge transfer in pre-trained models, such as fine-tuning \cite{howard2018universal,kumar2022fine,wortsman2022robust} and linear probing. Yet, they necessitate parameters or layer modifications, making them computationally intensive and less generalizable. 
To mitigate these issues, an efficient alternative known as Visual Prompting (VP) \cite{bahng2022exploring} or model reprogramming \cite{tsai2020transfer,elsayed2018adversarial,chen2021adversarial,neekhara2022cross} has emerged. It keeps the pre-trained model frozen while learning to add prompt to the inputs.
Since no changes have been made to the model itself and the prompts are with very few parameters, it achieves efficient and lightweight knowledge transfer.

\begin{figure}[t]
  \centering
   \includegraphics[width=0.48\textwidth]{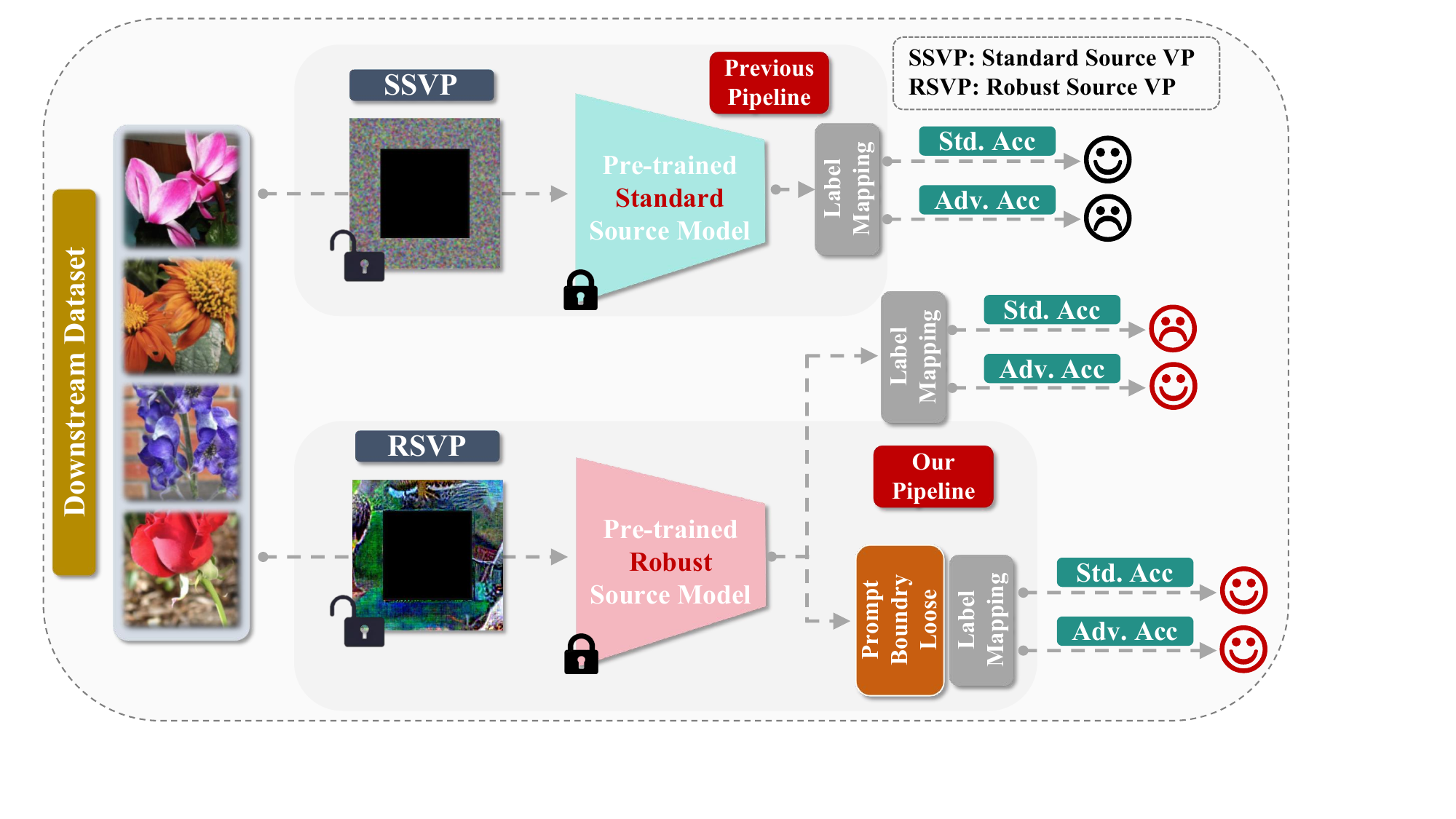}

   \caption{RSVP are visually more human aligned. The proposed PBL method under RSVP brings benefits both in robustness and generalization ability.}
      \vspace{-4mm}
   \label{fig:teaser}
\end{figure}
However, as shown in Fig.\ref{fig:teaser}, current works default set the source model as standard models obtained by standard training which is easy to be disturbed by adversarial attacks \cite{goodfellow2014explaining,chakraborty2018adversarial,zhang2019adversarial,ilyas2018black,guo2019simple,andriushchenko2020square,brendel2017decision,cheng2019improving,chen2020rays}. Correspondingly, robust models obtained by adversarial training \cite{shafahi2019adversarial,andriushchenko2020understanding,ganin2016domain,tramer2017ensemble,tramer2019adversarial,kurakin2016adversarial} have robustness (adversarial accuracy) against adversarial attacks but is often plagued by the decline of performance (standard accuracy) \cite{chan2019jacobian,tsipras2018robustness,gowal2020uncovering,xie2020adversarial,allen2022feature,zhang2019interpreting}, and the process of adversarial training requires much more computing resources \cite{wong2020fast,wang2019bilateral,jia2022boosting} than standard training due to its bi-level training process. Considering the good generalization ability of the VP from a Standard Source model (SSVP) and its lightweight in training, it is meaningful to study the properties of VP from a Robust Source model (RSVP). We naturally raise the following series of questions: Whether RSVP can inherit the robustness? Will it also suffer from suboptimal performance? If so, how to explain this phenomenon and how to alleviate it?

The gaps in comprehending the characteristics of RSVP and the absence of a definitive remedy for its defects spurred the advancement of our work, which, to the best of our knowledge, is the first work under this scenario. We find that RSVP can inherit the robustness of its source model, and also encounter the same sub-optimal standard accuracy that affects generalization ability. Besides, an explanation for the above phenomenon from the view of VP's visual representation has been proposed: RSVP is more in accordance with human perception. Moreover, we propose Prompt Boundary Loose (PBL), the first solution that aims at improving the generalization ability of RSVP. By maintaining the complex decision boundary of the robust model while increasing the mapping range of each label in a target downstream dataset, PBL successfully helps RSVP to maintain (or even greatly enhance) its robustness while redeeming its generalization ability. 

\textbf{Contribution.} We explored the previously uncharted territory of VP under an RSVP scenario. Our findings both quantitatively and qualitatively verify the inheritance of robustness from a robust source model in the VP tasks. We also proposed a strategy PBL to \textit{kill two birds with one stone}: Alleviate the generalization suboptimality of RSVP without negative (or even with significant positive) effects on robustness. Sufficient experiments fully prove the universality of the above phenomenon and the wide applicability of PBL.

\section{Related Work}
\label{sec:RelatedWork}
{\bf Prompt Learning in vision tasks.}
Given the success of prompt tuning in Natural language processing (NLP) \cite{brown2020language,devlin2018bert,liu2023pre,li2021prefix,lester2021power}, numerous studies have been proposed to explore its potential in other domains, such as vision-related and multi-modal scenarios \cite{chen2022adaptformer,gao2023clip,zhou2022conditional,zhou2022learning}. Nevertheless, most of these works still primarily focus on text prompting. 
VPT \cite{jia2022visual} takes the first step to visual prompting by adapting vision transformers to downstream tasks with a set of learnable tokens at the model input. Concurrently, VP \cite{bahng2022exploring} follows a pixel-level perspective to optimize task-specific patches that are incorporated with input images. Although not outperforming full fine-tuning, VP yields an advantage of parameter-efficiency, 
necessitating significantly fewer parameters and a smaller dataset to converge.

Subsequent works explored the properties of VP from different angles. \cite{chen2023understanding} proposed to use different label mapping methods to further tap the potential of VP. \cite{oh2023blackvip} proposed to restrict access to the structure and parameters of the pre-trained model, and put forward an effective scheme for learning VP under a more realistic setting. In addition, \cite{chen2023visual} explores the use of VP as a means of adversarial training to improve the robustness of the model, however, their method is limited to the in-domain setting, which is contrary to the original cross-domain transfer intention of VP. It is worth noting that current works on VP are all focused on scenarios where the pre-trained source model is a standard model, and no work has yet investigated the characteristics of VP when originating from a robust source model.

\noindent{\bf Robust Model and Adversarial Training.}
\cite{goodfellow2014explaining} were the first to propose the concept of adversarial examples, in which they added imperceptible perturbations to original samples, fooling the most advanced Deep Neural Networks (DNNs) of that time. Since then, an arms race of attack and defense has begun \cite{chakraborty2018adversarial,zhang2019adversarial,ilyas2018black,guo2019simple,andriushchenko2020square,jang2019adversarial,liao2018defense}, with numerous studies exploring different setups for attacks and defenses. Among the array of defense techniques, adversarial training stands out as the quintessential heuristic method
and has spawned a range of variant techniques \cite{tramer2017ensemble,xie2020smooth,ganin2016domain,shafahi2019adversarial,andriushchenko2020understanding,tramer2019adversarial,kurakin2016adversarial}. 
It's broadly recognized that although robust models may exhibit adversarial robustness, this typically comes at the expense of reduced standard accuracy\cite{chan2019jacobian,tsipras2018robustness,papernot2017practical,gowal2020uncovering,xie2020adversarial,allen2022feature,geirhos2018imagenet,zhang2019interpreting}, meaning their generalization abilities are compromised. 

Numerous studies have delved into the above trade-off phenomenon. \cite{tsipras2018robustness} proposed that there may exist an inherent tension between the goal of adversarial robustness and that of standard generalization, discovering that this phenomenon is a consequence of robust classifiers learning fundamentally different feature representations than standard classifiers. \cite{allen2022feature} analyzed the characteristics of adversarial training from the perspective of mixed features, pointing out that adversarial training could guide models to remove mixed features, leading to purified features (Feature Purification), thus visually conforming more to human perception. Moreover, some works believe that the trade-off between adversarial robustness and standard accuracy can be avoided \cite{pang2022robustness} and provide experimental or theoretical proofs. There is yet a perfect explanation for this phenomenon. Current research indicates that VP is effective at learning and transferring knowledge from standard source models.
However, the inheritance of the unique properties of robust source models by VP remains an area that urgently requires exploration. In this paper, we explored this hitherto unexplored territory for the first time and present the first solution to the negative effects observed in this scenario.

\section{Preliminaries}
{\bf Standard and Adversarial Training.}
In standard classification tasks, the main goal is to enhance standard accuracy, focusing on a model’s ability to generalize to new data samples that come from the same underlying distribution. The aim here is defined as achieving the lowest possible expected loss:

\begin{equation}\label{eq:1}
\min_{\theta} \mathbb{E}_{(x,y) \sim D} [\mathcal{L}(x,\theta,y)]
\end{equation}

where $(x,y) \sim D$ represents the training data $x$ and its label $y$ sampled from a particular underlying distribution $D$, 
and $\mathcal{L}$ represents the cross-entropy loss.

After \cite{goodfellow2014explaining} firstly introduce the concept of adversarial training, some subsequent works further refined this notion by formulating a min-max problem where the goal is to minimize classification errors against an adversary that add perturbations to the input to maximize these errors:

\begin{equation}\label{eq:2}
\min_{\theta} \mathbb{E}_{(x,y) \sim D} [\max_{\delta \in \Delta }\mathcal{L}(x+\delta,\theta,y)]
\end{equation}

where $\Delta$ refers to the set representing the perturbations allowed to be added to the training data $x$ within the maximum perturbation range $\epsilon$, we can define it as a set of $l_{p}$-bounded perturbation, i.e. $\Delta = \{ \delta \in R^{d} \mid \lVert \delta \rVert_{p} \leq \epsilon
  \}$. 

\noindent{\bf Visual Prompt Learning under Robust Model.} 
For a specific downstream dataset, the goal of visual prompt learning is to learn a prompt that can be added to the data thus allowing the knowledge of a pre-trained model to be transferred to it. The objective can be formally expressed as follows: 
\begin{equation}\label{eq:3}
\begin{aligned}
& \min_{\varphi} \, \mathbb{E}_{(x_{t},y_{t}) \sim D_{t}} [\mathcal{L}(\mathcal{M}(f_{\theta^{*}}(\gamma_{\varphi}(x_{t})), y_{t}))]  \\
& \text{s.t.} \quad \theta^{*} = \min_{\theta} \, \mathbb{E}_{(x_{s},y_{s}) \sim D_{s}} [\mathcal{L}(x_{s},\theta,y_{s})]
\end{aligned}
\end{equation}

when the pre-trained model is a robust model, the conditional term in Eq.\ref{eq:3} is changed to:
\begin{equation}\label{eq:4}
\theta^{*} = \min_{\theta} \, \mathbb{E}_{(x_{s},y_{s}) \sim D_{s}} [\max_{\delta \in \Delta }\mathcal{L}(x_{s}+\delta,\theta,y_{s})]
\end{equation}

where $D_{t}$ and $D_{s}$ represent the distribution of the downstream dataset and the source dataset, respectively; $f_{\theta^{*}}(\cdot)$ represents the frozen pre-trained model, which is parameterized by the optimal parameters $\theta^{*}$; $\gamma_{\varphi}({\cdot})$, parameterized by $\varphi$, represents the visual prompt that needs to be learned; $\mathcal{M}(\cdot)$ represents the pre-defined label mapping method.

\section{Method}
As mentioned earlier, 
existing works primarily focus on understanding VP in the context of standard models, the unique characteristics of inheritance under RSVP as well as solutions for its specific disadvantages remain to be explored. In this section, we address the previously raised questions and present our solution.

\subsection{Observations}
\begin{figure}[t]
\vspace{-3mm}
    \centering
    \begin{subfigure}[b]{\columnwidth}  
        \includegraphics[width=\textwidth]{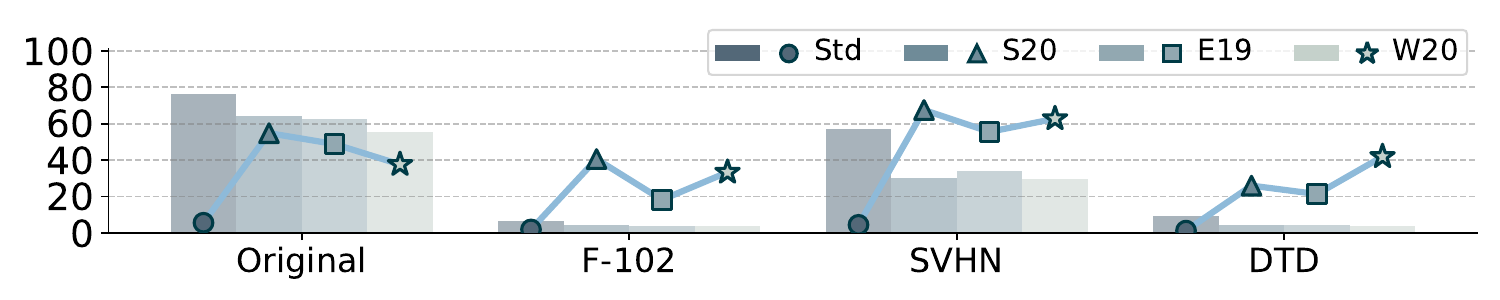}
        \caption{Random Label Mapping}
        \label{fig:sub1}
    \end{subfigure}

    \begin{subfigure}[b]{\columnwidth}  
        \includegraphics[width=\textwidth]{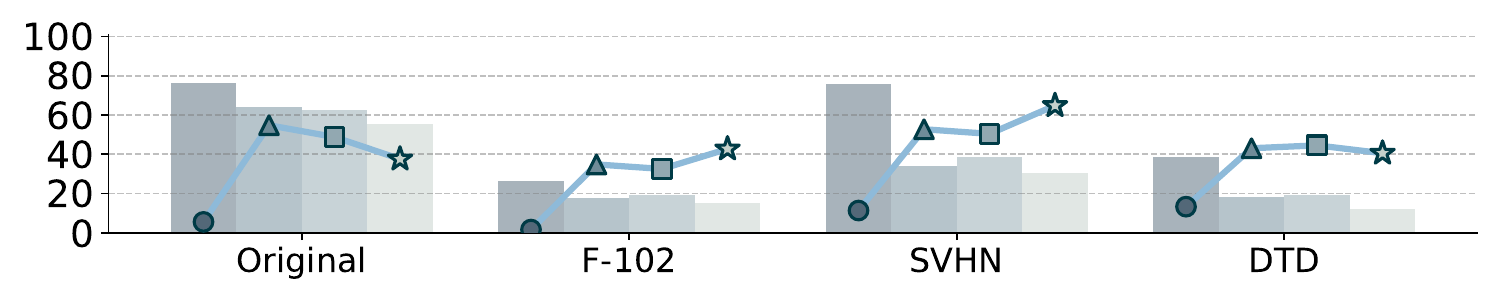}
        \caption{Iterative Label Mapping}
        \label{fig:sub2}
    \end{subfigure}

    \caption{The performance of VP on standard accuracy (histogram) and adversarial accuracy (line chart) when using a standard model or different robust models as the source model. `Original' represents the result on the source dataset without VP.}
    \label{fig:contrast}
    \vspace{-3mm}
\end{figure}

\noindent{\bf Robustness Inheritance of Visual Prompt.} Initially, we investigate the extent to which a source model's robustness transfers to visual prompts that are trained on downstream datasets distinct from the source dataset.
For the selection of robust models, we use the models collected in RobustBench \cite{croce2021robustbench}, an open-source benchmark which is widely used in the field of trustworthy machine learning. Specifically, we select one standard model and three robust models trained with ImageNet \cite{deng2009imagenet} under the $l_{\infty}$-norm, they are referred to as S20 \cite{salman2020adversarially}, E19 \cite{robustness} and W20 \cite{wong2020fast}. Without loss of generality, we used FGSM (Fast Gradient Sign Method) attack \cite{goodfellow2014explaining} to assess the model's robustness, datasets selected here are flowers102 (F-102) \cite{nilsback2008automated}, SVHN \cite{netzer2011reading} and DTD \cite{cimpoi2014describing}. The experiment results are shown in Fig.\ref{fig:contrast}, among which Fig.\ref{fig:contrast} (a) and Fig.\ref{fig:contrast} (b) represent the results under different label mapping methods, respectively. The bar chart represents the result of standard accuracy while the line chart represents the result of adversarial accuracy. `Original' represents the corresponding results of the source model on its original source dataset without VP.

\begin{figure}[t]
  \centering
   \includegraphics[width=0.48\textwidth]{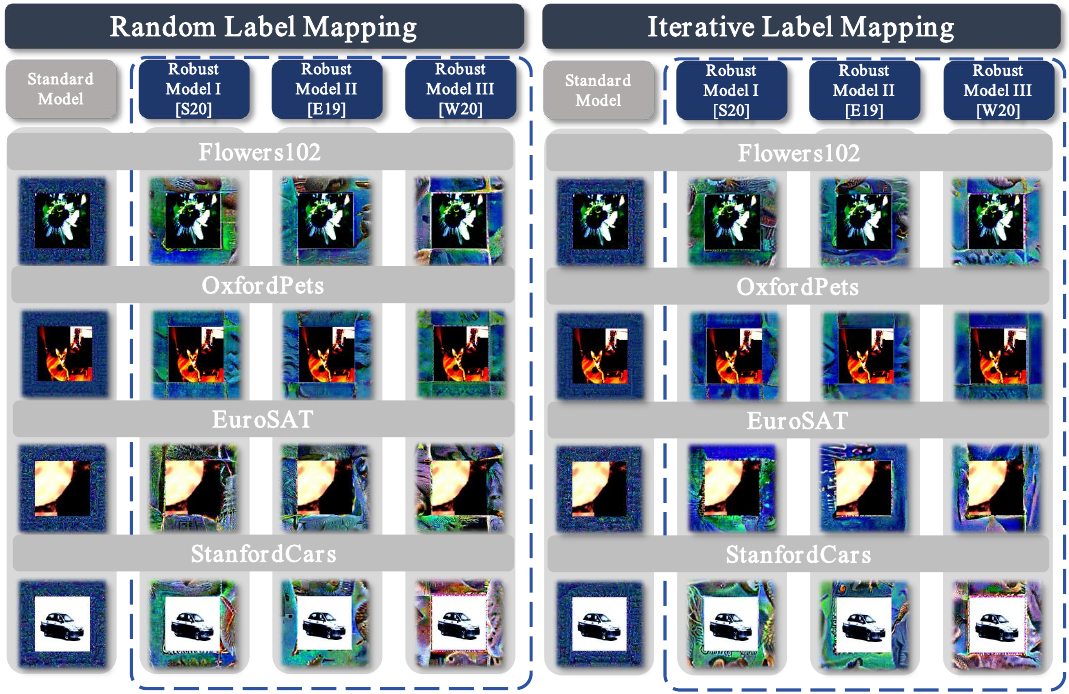}

   \caption{Visual representation of SSVP (columns 1 \& 5) and RSVP (columns 2-4 \& 6-8) obtained during a certain training period. For SSVP, only meaningless noise can be observed, while for RSVP, we get a representation consistent with human perception.}
      \vspace{-6mm}
   \label{fig:rbt_vp}
\end{figure}
\begin{figure*}[t]
  \centering
   \includegraphics[width=\textwidth]{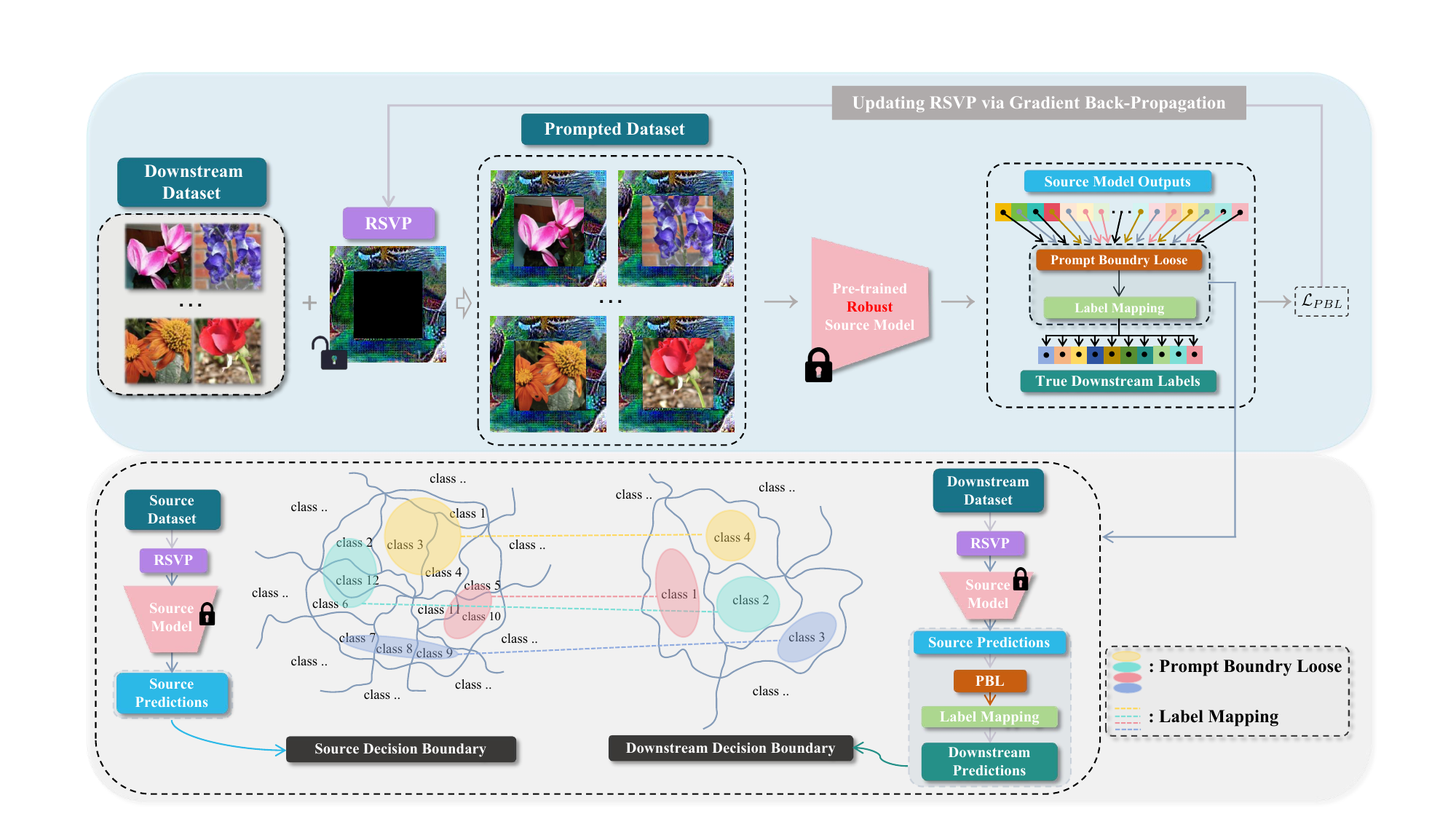}

   \caption{Pipeline of the proposed Prompt Boundary Loose (PBL).}
   \label{fig:ppl}
   \vspace{-0.1in}
\end{figure*}

The bar charts in Fig.\ref{fig:contrast} illustrate that visual prompts derived from a standard source model exhibit no robustness. In contrast, visual prompts trained with robust source models demonstrate markedly improved robustness compared to their standard-trained counterparts. Moreover, we observe that a given source model yields varying outcomes across different downstream datasets. Similarly, for a specific downstream dataset, the results differ when using various source models. 

\noindent{\bf Generalization Ability Encountered Degradation.} 
We further examine the disparities in standard accuracy between SSVP and RSVP across various downstream datasets. The line charts in Fig.\ref{fig:contrast} illustrate a decrease in generalization ability for RSVP compared to SSVP, mirroring the performance trend observed in the source model itself.

In addition, we can also find that there's no obvious relationship between the performance gaps of various robust source models and the RSVP performance disparities derived from them. This suggests that enhancing the robustness or generalization ability of the source model does not necessarily translate to similar improvements in RSVP. In fact, such attempts may be ineffective or even detrimental. Thus, a tailored approach is essential for RSVP to boost its generalization ability while maintaining or potentially increasing its robustness. Our proposed PBL method represents an initial foray into addressing this challenge.

\noindent{\bf Visual Representation of Visual Prompt from Robust Models.}
All current VP-related works focus on the case of SSVP. Under this setting, the resulting VP appears to be random noise without any meaningful visual representation, as shown in columns 1 and 5 of Fig.\ref{fig:rbt_vp}. In this work, we visualize RSVP and surprisingly find that RSVP (as shown in columns 2-4 and 5-8 of Fig.\ref{fig:rbt_vp}) can get a visual representation which aligns well with human perception. This phenomenon universally occurs across different robust models, different label mapping methods and different datasets.

The above phenomenon offers potential insights into RSVP's inheritance of robustness from source models. Recalling Eq.\ref{eq:3} and Eq.\ref{eq:4}, VP with learnable parameters receives as input an original image and gets an image-like output (hereinafter referred to as trainable image). This trainable image is then fed into the pre-trained source model for prediction. If VP and the original image are regarded as a whole, then the process of training VP can essentially be seen as the process of calculating and updating loss gradient with respect to part of the input image pixels. Prior works \cite{tsipras2018robustness,allen2022feature} suggest that adversarial robustness and standard generalization performance might be at odds with one another, which is attributed to the fact that the feature representations of standard and robust models are fundamentally different. To illustrate, in the absence of a Visual Prompt (VP), when one calculates the loss gradient with respect to the input image pixels (this operation can highlight the input features that significantly influence the loss and hence the model's prediction), it becomes evident upon visualization that robust models develop representations that are more aligned with prominent data features and human perception, which is consistent with the traits exhibited by RSVP.

\subsection{Prompt Boundary Loose}
The above experimental findings indicate that while RSVP does inherit the source model's robustness, it similarly suffers from a decline in standard accuracy akin to that of the source model, significantly constraining its practical applicability. As shown in Fig.\ref{fig:ppl}, we introduce the Prompt Boundary Loss (PBL) to solve this defect.

Previous research \cite{chen2023visual} delved into VP as a means of adversarial defense for a standard pre-trained model. Their findings indicated that while VP-enriched adversarial training can enhance model robustness, it also leads to a marked decrease in standard accuracy. 
The experiments conducted by \cite{chen2023visual} were confined to in-domain scenarios, i.e., the source and downstream datasets were identical. As shown later, we carry out experiments on further adversarial training of RSVP under cross-domain conditions and affirm that the observed trade-off between standard accuracy and robustness hold true as well.
To sum up, since RSVP itself suffers from the decline of standard accuracy, further adversarial training would only exacerbate this issue and increase computing resource consumption and time usage, which is unrealistic and meaningless.

Referring back to Eq.\ref{eq:3} and Eq.\ref{eq:4}, every input image from the target downstream dataset is first processed by RSVP, and then by the source model, to yield a predicted probability $f_{\theta^{*}}(\gamma_{\varphi}(x_{t}))$ that aligns dimensionally with that of the source dataset. Subsequently, by employing the pre-defined label mapping method $\mathcal{M}(\cdot)$,  we obtain the final predicted probability for the target dataset. 
In the RSVP scenario, the source model is an adversarial-trained model, whose decision boundary is more complex than a standard-trained one, while utilizing the pipeline of VP, the decision boundary of the frozen source model is unchangeable, which will greatly increase the learning difficulty of RSVP.
One might assume that enhancing the RSVP's learning capabilities from a complex decision boundary could be achieved by scaling it up to introduce more trainable parameters. Nevertheless, existing research \cite{bahng2022exploring} suggests that such scaling offers marginal benefits to VP performance, and beyond a certain threshold, it may even adversely affect its efficacy.
Motivated by the aforementioned insights and observations, we introduce PBL as an initial step towards advancing the functionality of RSVP. 

Specifically, PBL can be defined as a function $\mathcal{Q}(\cdot)$, which receives the output of the source model $f_{\theta^{*}}(\gamma_{\varphi}(x_{t}))$ and a temperature $\mathcal{T}$ as inputs, and combines the elements of $f_{\theta^{*}}(\gamma_{\varphi}(x_{t}))$ according to $\mathcal{T}$ to output an intermediate vector with a smaller dimension than the original output, then do the label mapping step $\mathcal{M}(\cdot)$ on this vector to get the final prediction for the target downstream dataset. By formalizing the objective function with PBL, we get:

\begin{equation}\label{eq:5}
\begin{aligned}
& \min_{\varphi} \, \mathbb{E}_{(x_{t},y_{t}) \sim D_{t}} [\mathcal{L_{PBL}}(\mathcal{M}(\mathcal{Q}(f_{\theta^{*}}(\gamma_{\varphi}(x_{t})), \mathcal{T}), y_{t}))] \\
& \text{s.t.} \quad \theta^{*} = \min_{\theta} \, \mathbb{E}_{(x_{s},y_{s}) \sim D_{s}}  [\max_{\delta \in \Delta }\mathcal{L}(x_{s}+\delta,\theta,y_{s})]
\end{aligned}
\end{equation}

We assume that the dimension of the output of the source model is $n$, and record the original output $f_{\theta^{*}}(\gamma_{\varphi}(x_{t}))$ as a vector $V = (v_1, v_2, ..., v_n)$. We deal with $n/\mathcal{T}$ elements at once and divide $V$ into $\mathcal{T}$ parts, each of which is marked as: 

\begin{equation}\label{eq:6}
\begin{aligned}
V_i = (v_{(i-1)n/\mathcal{T} + 1}, v_{(i-1)n/\mathcal{T} + 2}, ..., v_{in/\mathcal{T}}), & \\
i = 1, 2, ..., \mathcal{T}
\end{aligned}
\end{equation}

Suppose the intermediate vector is called $\mathcal{I}$, its $i^{th}$ element is the maximum value in the $i^{th}$ partition of $V$, i.e., $I_i = \max(V_i)$, which means taking the maximum confidence score in the current merged block as a representative value, as shown in Fig.\ref{fig:ppl}. $\mathcal{I}$ can be expressed as: 

\begin{equation}\label{eq:7}
\mathcal{I} = (\max(V_1), \max(V_2), ..., \max(V_T))
\end{equation}

The underlying intuition of the intermediate vector $\mathcal{I}$ lies in leveraging the knowledge the source model has learned from the source dataset to its fullest potential in the early stage of knowledge transfer. 
In addition, the looser decision area increases the quality of label mapping, thereby reducing the prediction difficulty of the downstream dataset. Finally, we can use $\mathcal{I}$ to map the downstream dataset and get final predictions:

\begin{equation}\label{eq:8}
\begin{aligned}
&\mathcal{L_{PBL}}(\mathcal{M}(\mathcal{Q}(f_{\theta^{*}}(\gamma_{\varphi}(x_{t})), \mathcal{T}), y_{t})) \\
& = \mathcal{L_{PBL}}(\mathcal{M}(\mathcal{Q}(V, \mathcal{T}), y_{t})) \\
& = \mathcal{L_{PBL}}(\mathcal{M}(\mathcal{I}, y_{t}))
\end{aligned}
\end{equation}

Note that upon applying VP to data from the same class in the downstream dataset, the source model may yield varying predictions (with the highest prediction probability associated with different classes). Furthermore, each data point within the same class could exhibit multiple high confidence scores.
The temperature $\mathcal{T}$ in PBL can formally loosen the decision boundary of $f_{\theta^{*}}(\cdot)$ thus reduce the difficulty of prediction and alleviate the low accuracy caused by the aforementioned phenomenon.
Simultaneously, it preserves and leverages the intricate decision boundary of the source model, ensuring that the robustness transferred from the source model is well conserved.
We find that PBL is highly compatible with existing label mapping methods; it can function as a seamless, plug-and-play enhancement to facilitate the training of a more effective VP.

\begin{table*}
\centering
\small
\setlength\tabcolsep{3pt} 
\resizebox{\textwidth}{!}{
\begin{tabular}{c|c|cc|cc|cc|cc|cc|cc}
\toprule
\toprule
\multirow{2}{*}{LM} & \multirow{2}{*}{Dataset} & \multicolumn{4}{c}{ResNet18} & \multicolumn{4}{c}{ResNet50} & \multicolumn{4}{c}{Wide-ResNet50-2} \\
\cmidrule(lr){3-6} \cmidrule(lr){7-10} \cmidrule(lr){11-14}
& & Adv. \footnotesize(w/o) & Adv. \footnotesize(w) & Std. \footnotesize(w/o) & Std. \footnotesize(w) & Adv. \footnotesize(w/o) & Adv. \footnotesize(w) & Std. \footnotesize(w/o) & Std. \footnotesize(w) & Adv. \footnotesize(w/o) & Adv. \footnotesize(w) & Std. \footnotesize(w/o) & Std. \footnotesize(w) \\
\midrule
\multirow{8}{*}{\rotatebox[origin=c]{90}{Random Label Mapping} } 
& F-102 & \textbf{33.33\%} & 32.79\% & 5.24\% & \textbf{7.43\%} & 40.57\% & \textbf{46.02\%} & 4.30\% & \textbf{4.63\%} & 19.33\% & \textbf{45.92\%} & 4.83\% & \textbf{5.24\%} \\
&DTD & 26.47\% & \textbf{39.36\%} & 4.02\% & \textbf{5.56\%} & 25.97\% & \textbf{37.17\%} & 4.55\% & \textbf{6.68\%} & \textbf{45.16\%} & 42.14\% & 5.50\% & \textbf{8.33\%} \\
&SVHN & 71.67\% & \textbf{74.21\%} & 32.23\% & \textbf{34.28\%} & \textbf{67.50\%} & 59.08\% & 30.18\% & \textbf{34.70\%} & 52.62\% & \textbf{58.16\%} & 35.50\% & \textbf{38.75\%} \\
&G-RB & 53.03\% & \textbf{78.71\%} & 12.42\% & \textbf{13.84\%} & 74.35\% & \textbf{77.16\%} & 11.95\% & \textbf{14.11\%} & 75.38\% & \textbf{80.99\%} & 15.08\% & \textbf{17.87\%} \\
&E-Sat & 46.45\% & \textbf{47.70\%} & 50.72\% & \textbf{53.46\%} & 43.59\% & \textbf{50.91\%} & 53.05\% & \textbf{57.78\%} & 42.46\% & \textbf{52.73\%} & 54.23\% & \textbf{56.31\%} \\
&O-Pets & 5.83\% & \textbf{14.75\%} & 3.27\% & \textbf{4.99\%} & 14.39\% & \textbf{16.57\%} & 3.60\% & \textbf{4.93\%} & 18.85\% & \textbf{24.60\%} & 3.33\% & \textbf{6.76\%} \\
&CI-100 & 75.07\% & \textbf{77.13\%} & 3.53\% & \textbf{4.95\%} & 66.26\% & \textbf{72.20\%} & 4.97\% & \textbf{5.16\%} & \textbf{77.77\%} & 74.06\% & 3.79\% & \textbf{5.61\%} \\
&S-Cars & 13.04\% & \textbf{13.11\%} & 0.57\% & \textbf{0.76\%} & 6.66\% & \textbf{20.37\%} & 0.56\% & \textbf{0.67\%} & 28.81\% & \textbf{32.69\%} & 0.73\% & \textbf{0.83\%} \\
\midrule
\multirow{8}{*}{\rotatebox[origin=c]{90}{Iterative Label Mapping} } 
& F-102 & 40.65\% & \textbf{44.66\%} & 18.88\% & \textbf{22.82\%} & 34.36\% & \textbf{34.86\%} & 17.70\% & \textbf{22.45\%} & 34.38\% & \textbf{37.25\%} & 19.53\% & \textbf{20.71\%} \\
& DTD & 41.11\% & \textbf{44.03\%} & 15.96\% & \textbf{18.79\%} & 43.09\% & \textbf{50.87\%} & 18.38\% & \textbf{20.45\%} & \textbf{54.28\%} & 50.14\% & 20.04\% & \textbf{21.45\%} \\
& SVHN & 61.67\% & \textbf{65.85\%} & 34.47\% & \textbf{35.44\%} & 52.76\% & \textbf{57.77\%} & 33.85\% & \textbf{34.96\%} & 52.85\% & \textbf{54.32\%} & 36.67\% & \textbf{37.60\%} \\
& G-RB & \textbf{68.96\%} & 67.92\% & 17.47\% & \textbf{20.24\%} & 74.42\% & \textbf{75.15\%} & 17.64\% & \textbf{19.46\%} & 62.23\% & \textbf{64.82\%} & 18.50\% & \textbf{19.26\%} \\
& E-Sat & 41.21\% & \textbf{42.13\%} & 59.20\% & \textbf{61.83\%} & 47.32\% & \textbf{47.36\%} & 58.12\% & \textbf{63.10\%} & \textbf{53.87\%} & 53.68\% & 55.59\% & \textbf{60.72\%} \\
& O-Pets & 32.84\% & \textbf{35.55\%} & 16.60\% & \textbf{23.00\%} & \textbf{38.53\%} & 38.15\% & 27.15\% & \textbf{33.74\%} & \textbf{38.25\%} & 37.18\% & 34.21\% & \textbf{36.17\%} \\
& CI-100 & 65.34\% & \textbf{68.99\%} & 11.60\% & \textbf{12.81\%} & \textbf{60.80\%} & 59.19\% & 11.51\% & \textbf{12.70\%} & 64.69\% & \textbf{64.71\%} & 10.97\% & \textbf{12.28\%} \\
& S-Cars & 20.26\% & \textbf{25.00\%} & 1.90\% & \textbf{2.29\%} & 24.44\% & \textbf{27.22\%} & 1.68\% & \textbf{2.10\%} & \textbf{33.57\%} & 33.14\% & 1.78\% & \textbf{2.23\%} \\
\bottomrule
\bottomrule
\end{tabular}}
\caption{Performance of our proposed Prompt Boundry Loose (PBL) under RSVP setting over eight downstream datasets and three pre-trained robust source models (ResNet-18, ResNet-50 and Wide-ResNet50-2 trained on ImageNet). Adv. (w/o) and Std. (w/o) means Adversarial Accuracy and Standard Accuracy without using PBL, while Adv. (w) and Std. (w) means Adversarial Accuracy and Standard Accuracy when using PBL. The better
outcomes are marked in bold.}
\label{tab:3}
\end{table*}
\section{Experiments}
In this section, we empirically demonstrate the effectiveness of the proposed PBL method in both adversarial and standard accuracy under RSVP on several different datasets and models, and explored the characteristics of PBL from multiple perspectives.

\begin{table}
\centering
\small
\setlength\tabcolsep{3pt} 
\resizebox{\linewidth}{!}{
\begin{tabular}{c|c|c|c|c|c|c|c|c}
\toprule
\toprule
Dataset & f-102 & DTD & SVHN & G-RB & E-Sat & O-Pets & Cl-100 & S-Cars \\ 
\midrule
$\mathcal{T}$       & 3     & 15  & 10   & 10   & 6     & 15     & 4      & 5      \\ 
\bottomrule
\bottomrule
\end{tabular}}
\caption{$\mathcal{T}$ used in different datasets.}
\vspace{-0.3in}
\label{tab:4}
\end{table}

\subsection{Experimental Settings}
\newcommand{\mysubsection}[1]{\noindent $\bullet$ \textbf{#1}}
\mysubsection{Models and Datasets.} We use two types of source model: Standard Source Model and Robust Source Model, both of which include ResNet-18, ResNet-50 and WideResNet-50-2 pre-trained on ImageNet-1K. For Standard Source Model, we use the pre-trained models from torch and timm \cite{rw2019timm}, while for Robust Source Model, we use the pre-trained models from RobustBench \cite{croce2021robustbench} same as in Fig.\ref{fig:contrast}, and for each backbone we select one robust model, all of which come from \cite{salman2020adversarially} and are collected by RobustBench. As for datasets, we consider to evaluate the performance of PBL over 8 downstream datasets: Flowers102 (F-102) \cite{nilsback2008automated}, DTD \cite{cimpoi2014describing}, GTSRB (G-RB) \cite{stallkamp2011german}, SVHN \cite{netzer2011reading}, EuroSAT (E-Sat) \cite{helber2018introducing, helber2019eurosat}, OxfordPets (O-Pets) \cite{parkhi2012cats}, StanfordCars (S-Cars) \cite{krause20133d} and CIFAR100 (CI-100) \cite{krizhevsky2009learning}.

\mysubsection{Evaluations and Baselines.}
Without lose of generality, we consider two different generally applicable label mapping methods \cite{chen2023understanding}: Random Label Mapping (RLM) and Iterative Label Mapping (ILM). RLM refers to randomly matching the labels of source dataset to those of the target dataset before training, while ILM refers to re-matching the labels of the source dataset to those of the target dataset according to the prediction frequencies of the source model after each iteration, so as to make full use of the training dynamics of VP. For each LM-Dataset-Backbone combination, we explore the standard accuracy (Std. Acc) as well as the adversarial accuracy (Adv. Acc) with or without PBL and FGSM is used as an attack method. 
Moreover, we delve into the impact of further adversarial training under RSVP, analyzing the results from the aspects of Std. Acc, Adv. Acc, time usage and computing resource consumption. Additionally, we dissect PBL's characteristics through various lenses, including temperature effects and prediction confidence dynamics.

\subsection{PBL brings benefits to RSVP}
Tab.\ref{tab:3} shows the effectiveness of our proposed PBL method under the RSVP scenario, considering the combination of 8 different datasets, 3 different backbones and 2 different LM methods. The first two columns for each backbone demonstrates the capability of  PBL in inheriting and maintaining (or even improving) robustness, while the latter two columns show its effectiveness in improving standard accuracy. The temperature $\mathcal{T}$ used under each dataset in Tab.\ref{tab:3} is shown in Tab.\ref{tab:4}.
It is significant to observe that a consistent temperature setting $\mathcal{T}$ across diverse backbone and LM method pairings yields uniform performance enhancements, which confirms the general advantages of PBL.
Subsequent experiments will detail how temperature $\mathcal{T}$ influences performance across various datasets.

Tab.\ref{tab:3} demonstrates that our proposed PBL method notably enhances performance in nearly all settings. Particularly, the standard accuracy achieved with PBL consistently surpasses that without it across all setups.
For example, with ResNet50 as backbone, Std. Acc of E-Sat dataset is improved by 4.73\% under RLM and 4.98\% under ILM. As for robustness, except for a slight decrease in a few setups, the Adv. Acc is well maintained or even greatly improved. For instance, when the backbone and LM methods are ResNet18 and RLM, the Adv. Acc of DTD increases by 12.89\% and the Adv. Acc of OxfordPets increases by 8.92\%. 
Moreover, our findings indicate that superior label mapping methods (e.g., ILM over RLM) can enhance standard accuracy but do not guarantee that VP can better inherit the robustness of the source model.
For instance, with ResNet18 as backbone and without PBL, robustness of E-Sat drops from 46.45\% under RLM to 41.21\% under ILM—a reduction of 5.24\%. Similarly, robustness of CI-100 decreases from 75.07\% with RLM to 65.34\% with ILM.
In most cases, our PBL method generally enables VP to better inherit robustness of the source model, regardless of the label mapping method applied.

\begin{figure}[t]
    \centering
    \hfill
    \begin{subfigure}[b]{0.44\textwidth}
        \includegraphics[width=\textwidth]{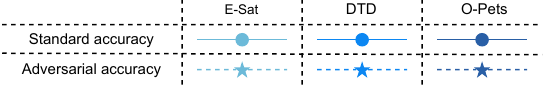}
    \end{subfigure}
    \begin{subfigure}[b]{0.24\textwidth}    
        \includegraphics[width=\textwidth]{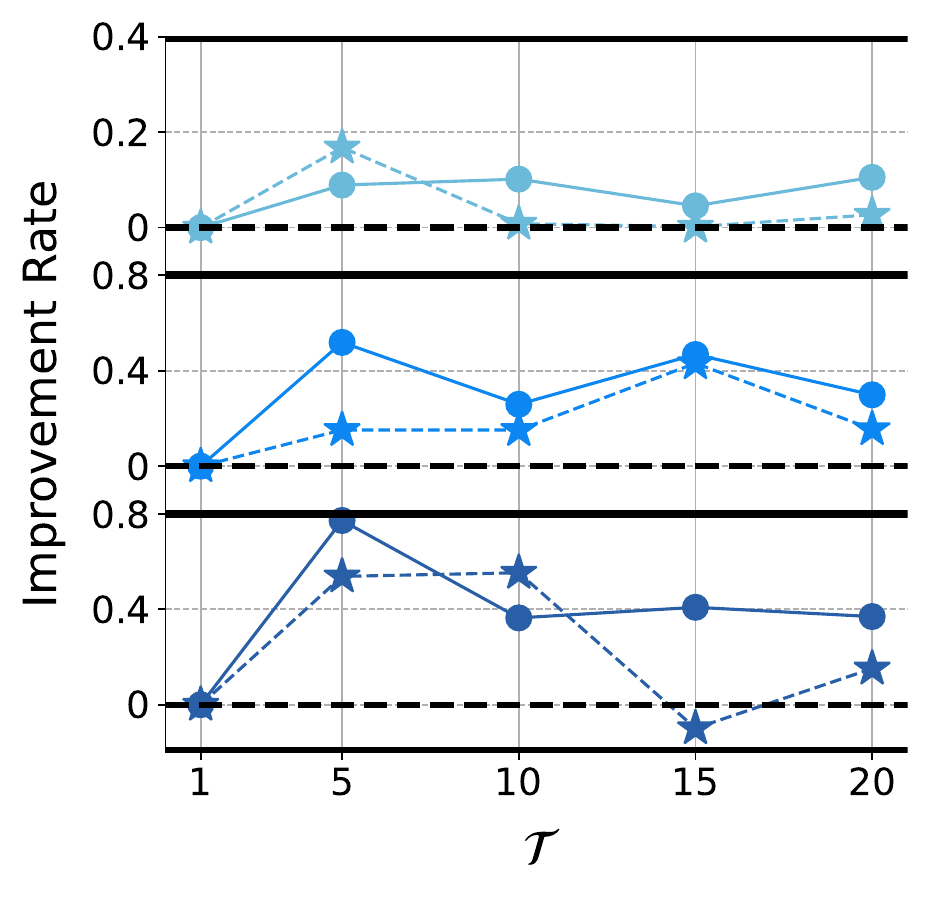}
        \caption{Random Mapping}
        \label{fig:random-mapping}
    \end{subfigure}
    \begin{subfigure}[b]{0.232\textwidth}
        \includegraphics[width=\textwidth]{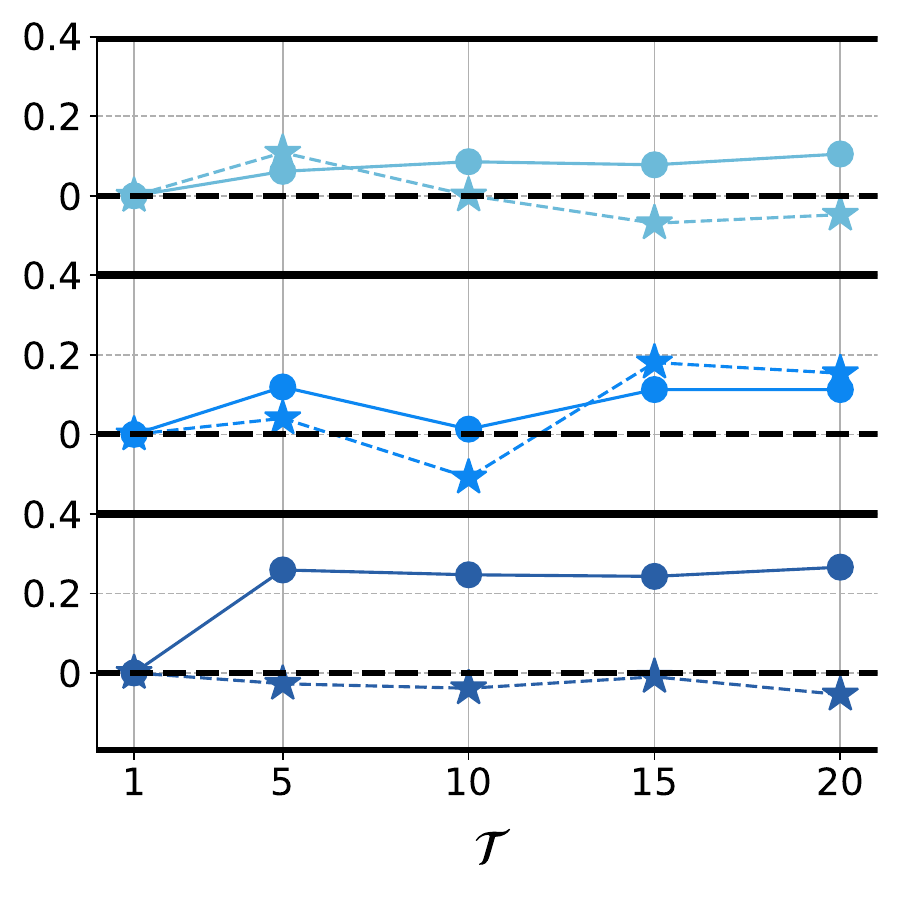}
        \caption{Iterative Mapping}
        \label{fig:iterative-mapping}
    \end{subfigure}

    \caption{The performance improvement of PBL in EuroSAT, DTD and OxfordPets at different temperatures $\mathcal{T}$, the standard accuracy is represented by solid lines and circles, while the adversarial accuracy is represented by dotted lines and asterisks. The range of temperature $\mathcal{T}$ is set between 1 and 20.}
    \label{fig:diff-T}
    \vspace{-2mm}
\end{figure}
Note that the computation of adversarial accuracy presupposes the model's correct initial classification of a sample—we only attempt an attack on samples that the model has accurately identified pre-attack.
Hence, employing PBL typically results in a larger set of samples subject to attack—attributable to the overall enhancement in standard accuracy. 
Therefore, when using PBL, it becomes more challenging to preserve or enhance the Adv. Acc of RSVP, thereby indicating that PBL provides a more accurate gauge of robustness.

\subsection{Understanding of PBL} 
In this section, we explore the characteristics of PBL and some potential factors that is likely to affect the performance.

\mysubsection{General advantages at different temperature $\mathcal{T}$.} We explored the effect of different temperature $\mathcal{T}$ on the performance of PBL. Without losing generality, we set temperature $\mathcal{T}$ to five values between 1 and 20 on EuroSAT, DTD and OxfordPets with ResNet50 as the backbone. The value of $\mathcal{T}$ = 1  is set as the zero point to indicate the baseline performance without PBL.  Performance at different temperatures is measured as the improvement rate relative to this baseline. The results are shown in Fig.\ref{fig:diff-T}. 

We can find that regardless of the temperature setting, PBL consistently yields substantial gains in standard accuracy across the board.
Specifically, PBL enhances standard accuracy by approximately 10\% across all temperature setups on EuroSAT. With DTD, employing RLM as the label mapping method typically results in a 40\% increase, while OxfordPets sees a peak improvement of around 80\%.
In addition, adversarial accuracy remains stable across various $\mathcal{T}$ values, with notable improvements observed at certain points. 
For example, with RLM, adversarial accuracy on E-SAT at $\mathcal{T}$ = 5, DTD at $\mathcal{T}$ = 15, and O-Pets at $\mathcal{T}$ = 5 increases by 20\%, 40\%, and 50\%, respectively.
\begin{figure}[t]
    \centering
    \begin{subfigure}[b]{0.9\columnwidth}    
        \includegraphics[width=\textwidth]{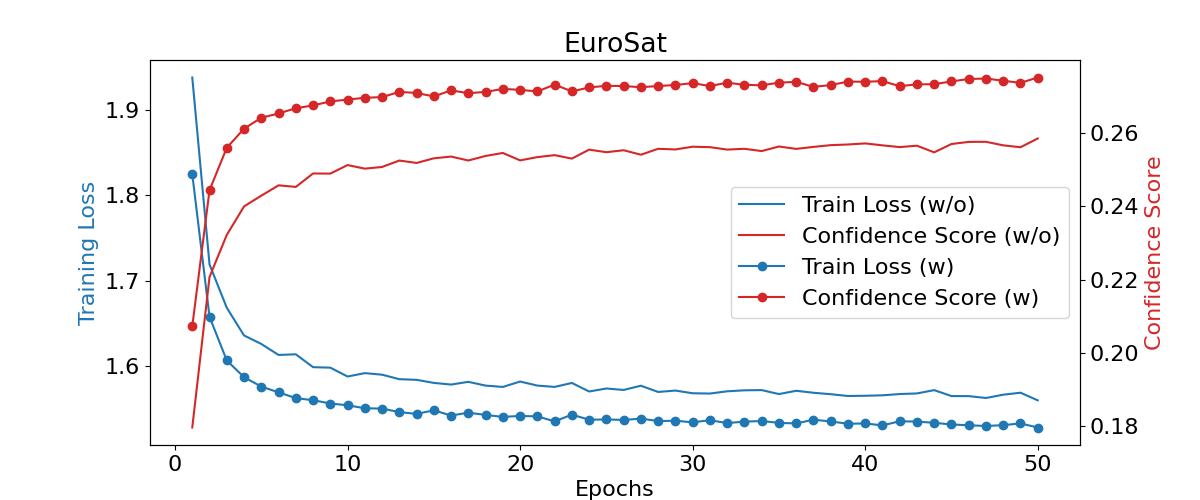}
        \caption{Training Dynamics of EuroSat}
        \label{fig:eurosat}
    \end{subfigure}
    \hfill
    \begin{subfigure}[b]{0.9\columnwidth}
        \includegraphics[width=\textwidth]{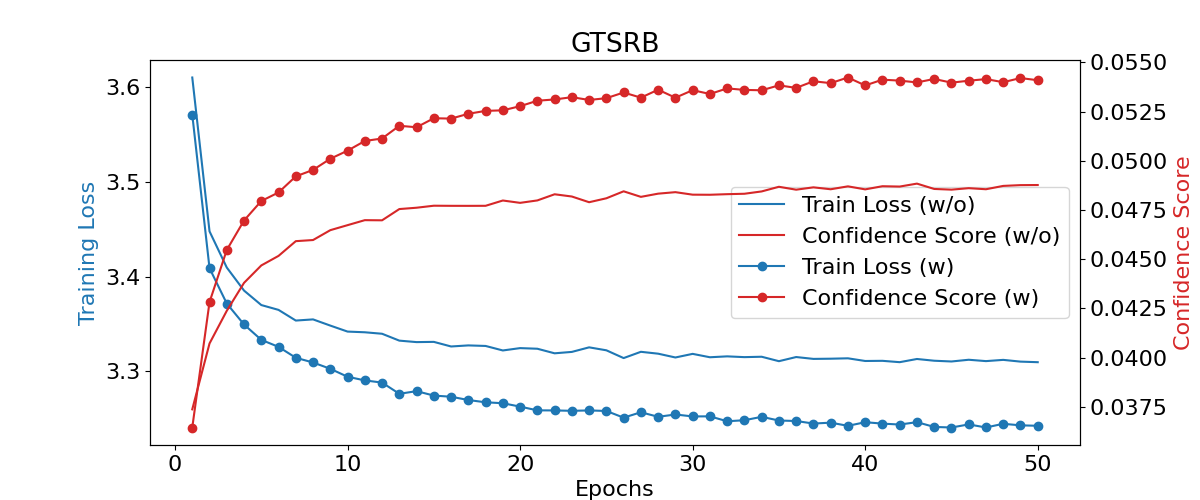}
        \caption{Training Dynamics of GTSRB}
        \label{fig:gtsrb}
    \end{subfigure}

    \caption{The training dynamics for the EuroSat and GTSRB datasets during the first 50 epochs utilizing RLM. PBL proves beneficial in the early stage of training.}
    \label{fig:dynamics}
    \vspace{-0.1in}
\end{figure}

\begin{figure*}[ht]
  \centering
  \begin{minipage}[t]{0.57\textwidth}
    \vspace{-0.3pt}
    \centering
    \small
    \resizebox{\linewidth}{!}{
    \begin{tabular}{c|c|c|c|c|c}
        \toprule
        \toprule
        Dataset    & Perf. & w/o. PBL & w/o. PBL+AT & w. PBL & w. PBL+AT \\
        \midrule
        \multirow{2}{*}{F-102} & Std. & 17.70\% & 16.16\% & 22.45\% & 19.20\% \\
        & Adv. & 34.36\% & 53.27\% & 34.86\% & 52.43\% \\
        \midrule
        \multirow{2}{*}{DTD} & Std. & 18.38\% & 17.61\% & 20.45\% & 19.27\% \\
        & Adv. & 43.09\% & 51.68\% & 50.87\% & 51.23\% \\
        \midrule
        \multirow{2}{*}{O-Pets} & Std. & 27.15\% & 24.83\% & 33.74\% & 31.53\% \\
        & Adv. & 38.53\% & 37.10\% & 38.15\% & 38.14\% \\
        \bottomrule
        \bottomrule
    \end{tabular}}
    \captionof{table}{The result of using four different combinations of strategies in different datasets. AT can improve robustness in some cases, however, sometimes it can not bring considerable gain but will consume more resources. In contrast, PBL can improve standard accuracy while maintaining robustness regardless of whether AT is utilized or not.}
    \label{tav:adv_train}
  \end{minipage}\hfill
  \begin{minipage}[t]{0.39\textwidth}
    \vspace{-0.3pt}
    \centering
    \includegraphics[width=\textwidth]{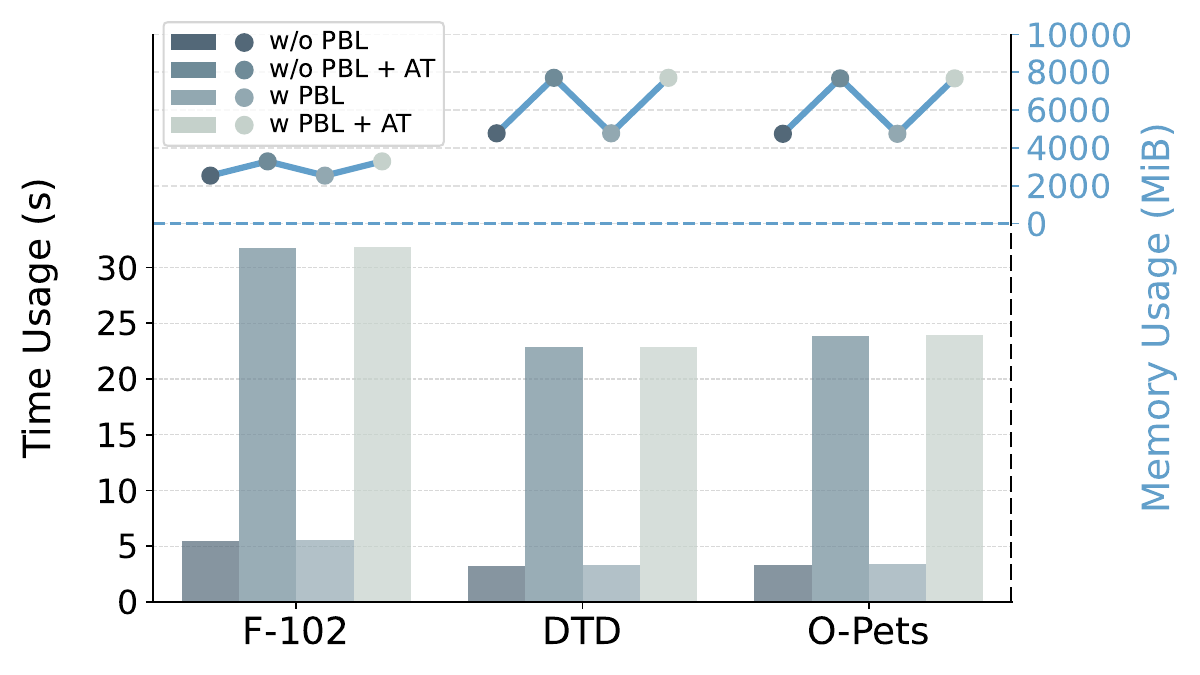}
    \captionof{figure}{Time usage and computing resource consumption under different combinations of PBL and AT. The bar chart represents time usage while the line chart represents the computing resource consumption. Results are the mean values per epoch.}
    \label{fig:adv_train}
  \end{minipage}
  \vspace{-0.5pt}
\end{figure*}
It is worth noting that different LM methods exhibit a consistent trend in standard accuracy gains across varying temperatures. For instance, on the DTD dataset, performance enhancement exhibits an `M-shaped' pattern with rising temperatures, peaking at $\mathcal{T}$ = 5 and $\mathcal{T}$ = 15, outperforming adjacent temperature values.
One possible explanation is that 
different LM methods may tap into specific phases of the VP training dynamics, including initialization and subsequent updates, to enhance overall performance.
RLM sets the mapping at beginning and maintains it throughout later iterations, making it dependent solely on the quality of the initialization.
ILM continuously revises its mapping sequence post-initialization (which can be seen as a re-initialization), capitalizing on the evolving training dynamics of VP.
Meanwhile, PBL select and pre-define a dynamic initialization for each training iteration from the potential distribution, enhancing the default settings and thereby improving the efficacy of different LM methods.
To verify this hypothesis, Fig.\ref{fig:dynamics} displays the training dynamics for the EuroSat and GTSRB datasets. From the onset, employing PBL yields a higher initial average confidence score and 
a lower training loss compared to the non-PBL setup. This advantage was maintained or even enhanced throughout the subsequent training process, demonstrating the superior performance offered by PBL in the initialization phase.

\begin{table}[h!]
  \centering
  \resizebox{\columnwidth}{!}{
  \begin{tabular}{c|c|cc|cc}
    \toprule
    \toprule
    \multirow{2}{*}{LM} & \multirow{2}{*}{Dataset} & \multicolumn{2}{c}{ResNet18} & \multicolumn{2}{c}{ResNet50} \\
    \cmidrule(lr){3-4} \cmidrule(l){5-6}
    & & {Std. Acc (w/o)} & {Std. Acc (w)} & {Std. Acc (w/o)} & {Std. Acc (w)} \\
    \midrule
    \multirow{3}{*}{\rotatebox[origin=c]{90}{RLM}} & f-102 & 12.02\% & \textbf{13.28\%} & 9.83\% & \textbf{12.06\%} \\
    & gtsrb & 47.14\% & \textbf{49.05\%} & 45.67\% & \textbf{46.83\%} \\
    & C-100 & 9.95\% & \textbf{11.36\%} & 9.61\% & \textbf{10.82\%} \\
    \midrule
    \multirow{3}{*}{\rotatebox[origin=c]{90}{ILM}} & f-102 & 29.03\% & \textbf{30.82\%} & 26.23\% & \textbf{26.67\%} \\
    & gtsrb & 52.86\% & \textbf{54.22\%} & 53.94\% & \textbf{55.61\%} \\
    & C-100 & 25.08\% & \textbf{27.34\%} & 38.87\% & \textbf{40.50\%} \\
    \bottomrule
    \bottomrule
  \end{tabular}}
  \caption{Comparison of standard accuracy with and without PBL under SSVP.}
  \label{tab:ssvp}
\end{table}
\mysubsection{PBL brings benefits to SSVP.}
It's intolerable to observe an improvement in standard accuracy solely for RSVP if it coincides with a substantial decrease for SSVP, as such a scenario would severely limit the practicality of the proposed method. 
Thus, we conduct additional experiments to assess PBL's performance with SSVP and anticipate that PBL will not detrimentally impact the generalization performance. 
As shown in Tab.\ref{tab:ssvp}, we are gratified to find that PBL not only markedly enhances the standard and adversarial accuracy in the RSVP context but also boosts the standard accuracy under SSVP—a welcome additional benefit, albeit not the primary aim of PBL.
Thus, PBL emerges as a versatile technique for enhancing the performance of VP across various source model types.

\mysubsection{The intolerability of adversarial training for VP.}  
We further examine the efficacy of additional adversarial training for RSVP in cross-domain transfer learning context. 
Note that the standard accuracy for RSVP is considerably lower than that for SSVP as a price of robustness. Thus, additional adversarial training for RSVP could exacerbate the reduction in standard accuracy. While this may yield robustness, a model that is robust yet lacks generalization ability is meaningless.

In Tab.\ref{tav:adv_train} and Fig.\ref{fig:adv_train}, we assess the impact of PBL and Adversarial Training (AT). Our analysis encompasses standard and adversarial accuracy, as well as average time usage and computing resource consumption over 200 training epochs, under four distinct combinations of PBL and AT. 
As shown in Tab.\ref{tav:adv_train}, while adversarial training alone enhances RSVP's robustness (columns 1 \& 2), it notably compromises standard accuracy. 
Even in some cases, e.g, with DTD and OxfordPets as target datasets, adversarial training not only leads to a reduction in standard accuracy but also offers negligible robustness gains (columns 2 \& 3), while significantly increasing computational resource consumption ($ \simeq 1.5 \times $) and time usage ($ \simeq 6 \times $), which is intolerable.
In contrast, applying PBL without adversarial training (columns 1 \& 3) enhances the standard accuracy of RSVP and preserves or even boosts its robustness. When combining PBL with adversarial training, PBL mitigates the drop in standard accuracy typically induced by adversarial training and sustains robustness enhancements (columns 2 \& 4), without additional time usage or computational resource consumption.
\section{Conclusion}
In this paper, we undertake an unprecedented exploration of the properties of Robust Source Model Visual Prompt (RSVP). We discover that RSVP inherit the robustness of the source model and provide an interpretation at visual representation level. Moreover, RSVP also experience suboptimal results in terms of standard accuracy. To address this problem, we introduce the first solution known as Prompt Boundary Loose (PBL), aiming at reducing the learning difficulty of RSVP by formally relaxing the decision boundary of the source model in conjunction with various label mapping methods. Extensive experiments results demonstrate that our proposed PBL not only maintains the robustness of RSVP but also enhances its generalization ability for various downstream datasets. 

{
    \small
    \bibliographystyle{ieeenat_fullname}
    \bibliography{main}
}

\end{document}